\pdfoutput=1

\documentclass[11pt]{article}

\usepackage[]{acl}

\usepackage{times}
\usepackage{latexsym}

\usepackage[T1]{fontenc}

\usepackage[utf8]{inputenc}

\usepackage{microtype}

\usepackage{graphicx}
\usepackage{float}
\graphicspath{{./figures/}}
\usepackage{flafter} 

\usepackage{csvsimple}
\usepackage{booktabs}
\usepackage{caption} 
\usepackage{adjustbox}
\usepackage{graphics}
\usepackage{needspace}
\usepackage{textcomp}
\makeatletter
\csvset{
  autotabularcenter/.style={
    file=#1,
    after head=\csv@pretable\begin{tabular}{|*{\csv@columncount}{c|}}\csv@tablehead,
    table head=\hline\csvlinetotablerow\\\hline,
    late after line=\\,
    table foot=\\\hline,
    late after last line=\csv@tablefoot\end{tabular}\csv@posttable,
    command=\csvlinetotablerow},
  autobooktabularcenter/.style={
    file=#1,
    after head=\csv@pretable\begin{tabular}{*{\csv@columncount}{c}}\csv@tablehead,
    table head=\toprule\csvlinetotablerow\\\midrule,
    late after line=\\,
    table foot=\\\bottomrule,
    late after last line=\csv@tablefoot\end{tabular}\csv@posttable,
    command=\csvlinetotablerow},
}
\makeatother

\newcommand{\reffig}[1]{\figurename~{\ref{fig:#1}}}

\newcommand{\framework}{LOOVE}
\newcommand{\pseudodict}{emote pseudo-dictionary}
\newcommand{\pseudodictupper}{Emote Pseudo-Dictionary}
\newcommand{\wv}{w2v}
\newcommand{\clfaccuracy}{71.16\%}
\newcommand{\pointsahead}{7.36}

\newcommand{\luc}[1]{{\color{red} LUC #1}}
\newcommand{\pavel}[1]{{\color{blue} PAVEL #1}}
\newcommand{\andrea}[1]{{\color{purple} ANDREA #1}}
\newcommand{\gentodo}[1]{{\color{orange} TODO #1}}

\renewcommand{\luc}[1]{}
\renewcommand{\pavel}[1]{}
\renewcommand{\andrea}[1]{}
\renewcommand{\gentodo}[1]{}

\title{FeelsGoodMan: Inferring Semantics of Twitch Neologisms}

 \author{Pavel Dolin \quad Luc d'Hauthuille \quad Andrea Vattani \\
         Spiketrap \\ San Francisco, CA, USA \\  \texttt{\{pavel, luc, andrea\}@spiketrap.io}}

\begin{document}
\maketitle

\begin{abstract}
Twitch chat messages pose a unique problem in natural language understanding due to a large presence of neologisms, specifically emotes. There are a total of 8.06 million emotes, over 400k of which were observed during the study period. There is virtually no information on the meaning or sentiment of emotes, and with a constant influx of new emotes and drift in both their frequencies and their perceived meanings, it becomes impossible to maintain an updated manually-labeled dataset. Our paper makes a two-fold contribution. First, we establish a new baseline for sentiment analysis on Twitch data, outperforming the previous benchmark by \pointsahead{} percentage points. Secondly, we introduce a simple but powerful unsupervised framework based on word embeddings and $k$-NN to enrich existing models with out-of-vocabulary knowledge. 
This framework allows us to auto-generate an \pseudodict{}, and we show that we can nearly match the supervised benchmark above, even when injecting such emote knowledge into sentiment classifiers trained on extraneous datasets.
\end{abstract}

\section{Introduction}
Live streaming platforms such as Amazon Twitch or YouTube Live have become increasingly popular in the last decade and have seen an even faster growth in the last couple of years due to the COVID-19 pandemic and the rise of esports. Users on these platforms watch videogame players livestream their gameplay, and comment live on the stream to share their opinions with the streamer and the rest of the audience. Given the instantaneous and idiosyncratic nature of the chat room culture, the language used is very different from a formal conversation.  It is riddled with grammatical errors, abbreviations, game-specific lingo, as well as emoji and emoji-like icons. In particular, Twitch users make heavy usage of \emph {emotes}, which are Twitch-specific icons or animations used to express a particular emotion, feeling, or inside-joke\footnote{Even though there is a visual component to emotes (as well as emojis), we focus on the NLP understanding of Twitch messages, interpreting emotes as words. Extracting signals from their visual counterparts is outside of the scope of this manuscript.}.

Emotes on Twitch have become a language of their own and have both changed and enriched how people communicate with each other on the platform. They can be interspersed within text to change the meaning of a message, or constitute an entire message on their own. They are rendered when users type a predefined string, e.g. ``Kappa''~\textrightarrow~\includegraphics[height=\fontcharht\font`\B]{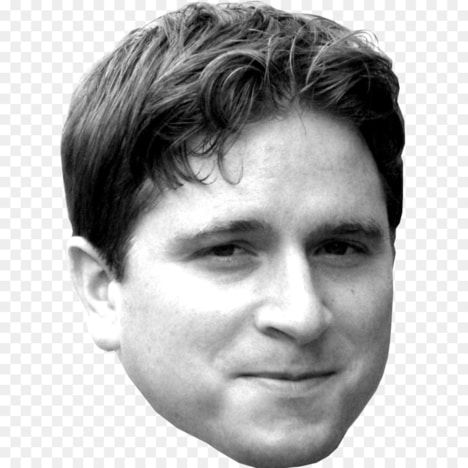} and ``LUL''~\textrightarrow~\includegraphics[height=\fontcharht\font`\B]{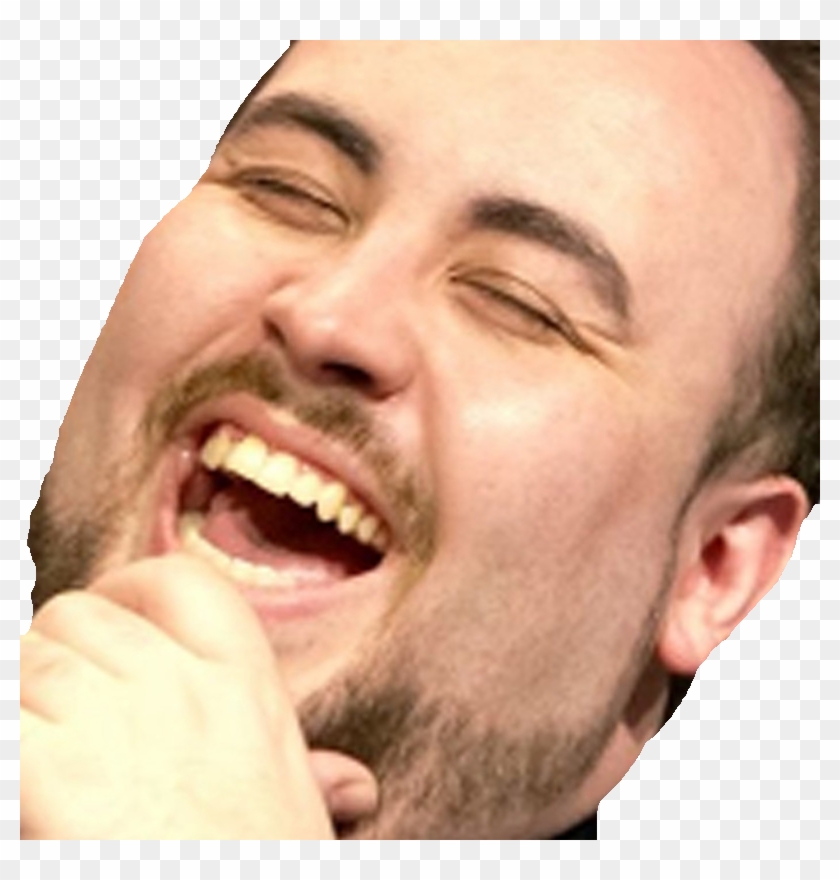}. There are over 8 million emotes---over 400,000 were observed in the week surveyed, constituting one third of all unique tokens on Twitch.  Like memes, emotes are generated by the community, causing a constant change in their frequency and meaning. 

One emote which whose meaning has changed over time is ``FeelsGoodMan''~\textrightarrow~\includegraphics[height=\fontcharht\font`\B]{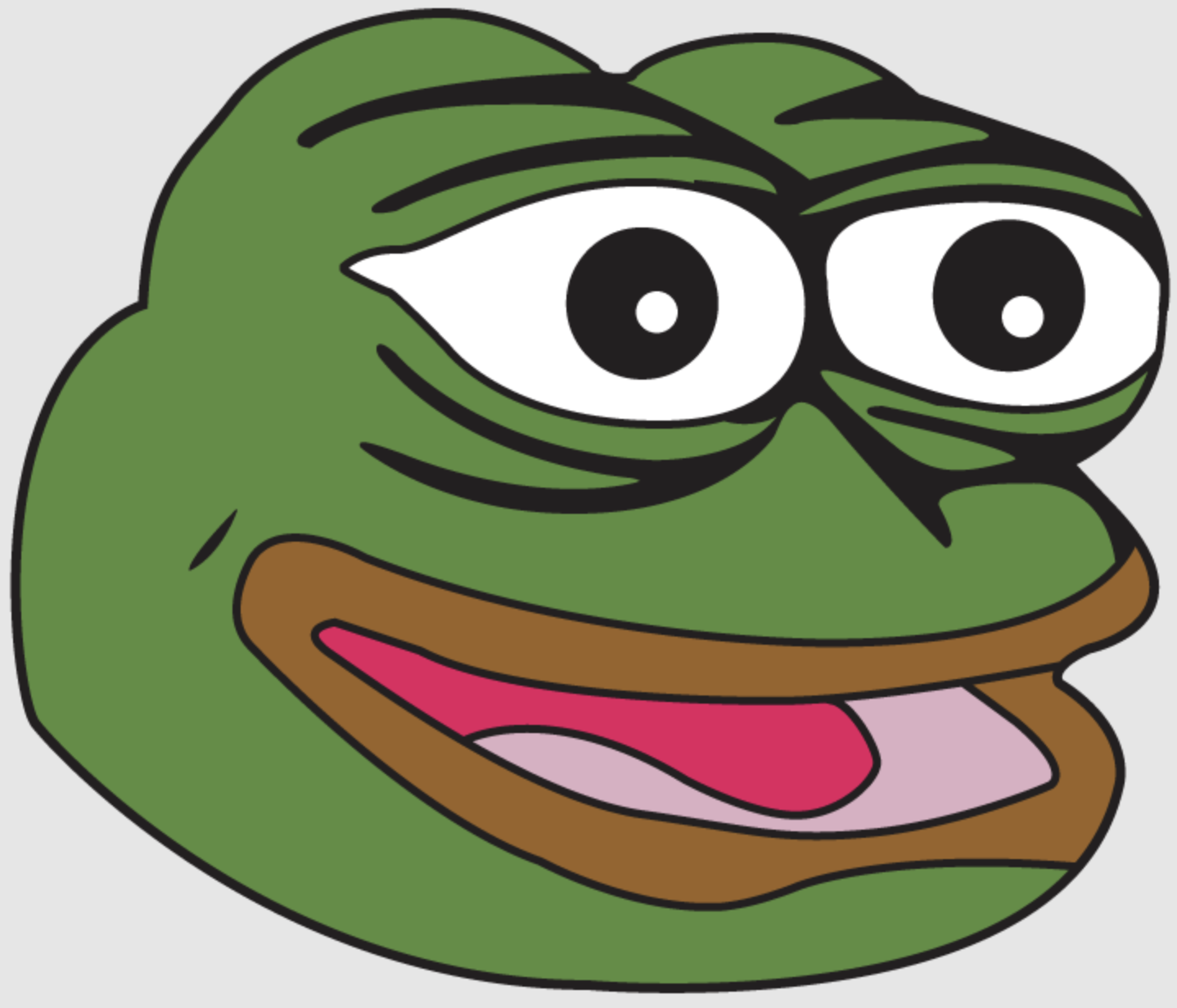}, based on a cartoon frog from a 2005 comic by the artist Matt Furie.  Furie's cartoon frog was adopted by right wing posters on various online forums like 4chan in the early 2010s.  Since then, Furie has campaigned to reclaim the meaning of his character, and the emote has seen an upsurge in more mainstream non hate usage \cite{Pepe2019} and positive usage on Twitch.  Our results on Twitch agree, showing that ``FeelsGoodMan'' and its counterpart ``FeelsBadMan'' are mainly being used literally.  

Continuous introduction of new emotes and their cryptic origins makes it unfeasible to maintain curated dictionaries documenting their meaning and semantics. With the exception of the recent work by \cite{kobs_emote-controlled_2020}, which classified 100 top emotes and labeled $2000$ Twitch chat messages, there is a lack of analytical studies focusing on understanding Twitch data and the enigmatic language of emotes. In this paper, we aim to fill this gap. 

\section{Our contribution}\label{sec:contribution}

In this paper we set to address two core tasks:
\begin{enumerate}
 	\item[(A)] Perform sentiment analysis on Twitch data better than previous baselines set by \cite{kobs_emote-controlled_2020}. In addition, we introduce a framework that can handle emote drift without additional major data labeling effort.
	\item[(B)] Provide a broad insight into emote semantics and their sentiment. This is to address the lack of a broad understanding of thousands of emotes.
\end{enumerate}

To address Task (A):
\begin{itemize}
    \item We conducted a thorough set of experiments comparing standard traditional machine learning methods for supervised sentiment analysis on Twitch data. To the best of our knowledge, no such foundational analyses have been performed; only a lexicon-based approach and a deep learning approach with noisy labels have been tried by \cite{kobs_emote-controlled_2020}.
    \item We show that our best model outperforms the previous benchmark set by \cite{kobs_emote-controlled_2020} by \pointsahead{} percentage points on accuracy.
    \item We break down the performance of our base classifiers and demonstrate that features with emotes constitute more than 50\% of feature importance, while comprising only 20\% of features.
	\item We introduce a drift-resilient framework to Learn Out Of Vocabulary Emotions (LOOVE). Requiring little to no additional data labeling, LOOVE is able to incorporate new emote knowledge into existing models without relying on emotes as explicit features.
\end{itemize}
\par

As for Task (B):
\begin{itemize}
    \item We create an \pseudodict{} based on word embedding neighborhoods.
    \item We automatically infer a corresponding sentiment for thousands of emotes from the \pseudodict{}.
\end{itemize}

The remainder of the paper is organized as follows: Section~\ref{sec:relatedwork} provides an overview of the related literature. Section~\ref{sec:baselines} presents our experiments to establish a new set of supervised baselines on the Twitch dataset. In Section~\ref{sec:framework}, we introduce our framework \framework{} to augment external classifiers with emote knowledge. Finally, Section~\ref{sec:discussion} discusses the construction and properties of the \pseudodict{}. 

Additionally, in the Appendix \ref{sec:appendix-emote_case_study} we present the Emote Case study. In Appendix \ref{sec:appendix-datrasets} we study of trends in the Twitch Unlabeled Dataset that we collected for the study. In Appendix \ref{sec:appendix-embeddings} we showcase additional applications of Twitch \wv{} embeddings.

\section{Related Work}\label{sec:relatedwork}
There are few studies on Twitch and emotes.  We relied on all existing relevant Twitch studies as well as the relevant literature found on other neologisms such as emoticons, emoji and slang.

\subsection{Emotes and Twitch} Labeled Twitch emote data is virtually non-existent. \citet{kobs_emote-controlled_2020} conducted a study with the Twitch community and semantically labeled 100 frequently occurring emotes in 2018. Although the top 100 emotes account for 35.1\% of the tokens and the top 1000 account for 52.1\% of tokens, there are over 8 million total emotes, with over 400,000 emotes observed in the week studied, and the number growing every day. The primary contribution of that work was providing the sentiment analysis baseline for Twitch data, as well as a labeled dataset of ~2000 chat messages. For the baseline, they used an Average Based Lexicon approach, which represents a comment as a sequence of tokens with an assigned sentiment from a look up table. To come up with the sentiment score for the entire comment, they averaged the sentiment scores of the tokens. This approach along with its variation achieved a 61.8\% and 62.8\% accuracy, respectively. They also employed a Convolutional Neural Network (CNN) approach, which was weakly-trained on the data generated by the Average Based Lexicon approach. This resulted in 63.8\% accuracy. 

Another noteworthy study of emotes was done by \cite{barbieri_towards_2017}, who tried to address emote prediction, i.e. which emote the user is more likely to use, and trolling detection, i.e. ``a specific set of emotes which are broadly used by Twitch users in troll messages.'' The authors were only predicting the top 30 most frequently used emotes. The highest F1 score for emote prediction was 0.39. The highest F1 score for trolling detection was 0.81.

Other emote and Twitch related studies include ``Classification of viewers by their consumption behavior and analysis of subscribers’ emote usage'' \cite{oh_cross-cultural_2020}; prediction of subscription status of a user in a channel based on user's comments \cite{loures_stinkycheese_2020}; and a master's thesis on language variety on Twitch entitled ``The present text is a research into the language usage in Computer-Mediated Communication, specifically on the online streaming platform Twitch.tv.'' \cite{hope_hello_2019}.

\subsection{Emoji and Emoticons} 
The sentiment analysis involving emojis and emoticons have been addressed with both deep learning (DL) and traditional machine learning (ML) approaches using various datasets. A notable DL approach is a emotion-semantic-enhanced bidirectional long short-term memory (BiLSTM) network with the multi-head attention mechanism model (EBILSTM-MH) \cite{wang_emotion-semantic-enhanced_2020}. This achieved 71.70\% accuracy on microblog text data involving emojis. Additionally authors showed that an SVM model achieved 66.81 accuracy on the same dataset. Another traditional ML approach for sentiment analysis, in this case using Twitter data, was carried out by \cite{illendula_learning_2018}. It is based on emoji embeddings from 147 million tweets, trained on Random Forest (RF) and  Support Vector Machine (SVM) with an overall accuracy of 62.1\% and 65.2\% respectively.  This approach outperformed the then current state of the art.

Studies have been done to understand the semantics of emoji and emoticons. One notable work is EmojiNet by \cite{wijeratne_emojinet_2016}. This is the first machine readable sense inventory for emojis. The researcher created a centralized table of emoji definitions, incorporated from multiple resources. Additionally, through word sense disambiguation techniques they assigned senses to emojis. Additionally, \cite{illendula_learning_2018} presented a thorough study of emoji semantics and their use in social media posts.

\subsection{Slang} 

\cite{wilson_urban_2020} used Urban Dictionary \footnote{\url{https://www.urbandictionary.com/}} as a corpus to create slang word embeddings. For sentiment prediction (64.4\% accuracy) and sarcasm prediction (80.2\% accuracy) they achieved marginally better scores than other standard pre-trained embeddings such as  word2vec-GoogleNews when initializing classifiers with UD embeddings. Other notable approaches include: a framework that combines probabilistic inference with neural contrastive learning that models the speaker's word choice in a slang context \cite{sun_computational_2021}. In addition a BiLSTM based model was utilized for slang detection and identification at the sentence level with an F1-score of 0.80.

\section{Supervised Sentiment on Twitch: new SoTA}\label{sec:baselines}

In this section we establish a new set of baselines for Twitch chat sentiment outperforming the past state of the art method \cite{kobs_emote-controlled_2020} by \pointsahead{} percentage points on accuracy. We also investigate the driving features behind the method, showing that emotes contribute significantly to the performance of the model, constituting on average over half of the Gini feature importance (using a Random Forest classifier). This is despite being only a fifth of the classifier's features.

\subsection{Dataset} For our training and testing corpus we used the Twitch sentiment dataset by \cite{kobs_emote-controlled_2020} which we will refer to as the Emote Controlled dataset or EC. This dataset is composed of $1,880$ examples with 40.6\%/38.0\%/21.4\% positive/neutral/negative class split. Data was split in a stratified fashion; designating 80\% of the data, 1502 examples for training and 20\%, 378 examples for testing \footnote{A 5-fold cross validation evaluation is consistent with the results obtained with a fixed split.}. 

\subsection{Features \& Models} 

We focused on Twitch chat sentiment analysis using traditional ML approaches, because, to the best of our knowledge, it had not previously been investigated. We trained Naive Bayes (NB), Logistic Regression or Maximum Entropy (ME), Random Forest (RF) and Support Vector Machines with linear kernel (SVM) models as they have been the most popular traditional ML algorithms for sentiment detection to date \cite{yadav_sentiment_2020, zimbra_state---art_2018}.

The input features to the models were constructed using a well established simple sentiment analysis approach based on a bag-of-features \cite{pang_thumbs_2002}. We tested both unigrams and unigrams plus bigrams as input features. We additionally tried three text processing methods. In our study we call minimal text processing Processing 1 (P1). It involves punctuation removal, lower-casing tokens and removing like characters that occur more than three consecutive times \footnote{For instance: \emph{loooove -> looove, haaaaaate -> hate.}}.  Processing 2 (P2) refers to P1 processing plus stop word removal. Processing 3 (P3) is P2 plus lemmatization of tokens.

\subsection{Results}
To our surprise, the ``textbook'' ML approaches with minimal processing, which we call P1, outperforms the previous Twitch sentiment baselines \cite{kobs_emote-controlled_2020} of 63.8\% by \pointsahead{} percentage points on accuracy in the best case on the same EC dataset. The accuracy results are summarized in Table~\ref{tab:ec-results-accuracy}.

The first column of Table~\ref{tab:ec-results-accuracy} describes the classifier and the type of input features, the rest of the columns are processing type. The integer following the classifier's name refers to the number of ngrams generated from the corpus \footnote{For example, NB.1 is a Naive Bayes classifier trained on unigrams, while RF.2 is a Random Forest classifier trained on unigrams and bigrams.}.

\begin{table}[ht]
\centering
\begin{adjustbox}{width=0.8\columnwidth,center}
\begin{filecontents*}{./csv/ec_process_x_classifier_x_ngram/accuracy.csv}
Clf,P1,P2,P3
NB.1,61.9,59.73,58.63
NB.2,60.58,59.45,58.9
ME.1,68.52,68.22,67.12
ME.2,69.31,69.59,68.49
SVM.1,67.72,69.59,69.32
SVM.2,68.78,69.59,68.22
RF.1,70.37,66.85,67.4
RF.2,\textbf{71.16},68.49,67.95
\end{filecontents*}
\csvautobooktabular{./csv/ec_process_x_classifier_x_ngram/accuracy.csv}
\end{adjustbox}
\caption{EC Test Dataset Accuracy for various models \label{tab:ec-results-accuracy}}
\end{table}

From Table \ref{tab:ec-results-accuracy}  it is evident that RF.2 with P1 processing, P1.RF.2, outperforms all benchmarks in accuracy delivering \clfaccuracy{}.

We break down the performance of P1.RF.2 to demonstrate the driving features behind the classifier. Table \ref{tab:sum-features-contributions} shows the cumulative sum of the Gini RF feature importance. Because we trained a ternary one-versus-rest classifier, results in the table are displayed for each class: positive, negative and neutral. Additionally, because the features consist of unigrams and bigrams, emotes can occur as unigrams and as a part of bigram. We differentiate it as follows:  the column ``emotes only'' refers to unigram emote features and the ``emotes+'' column refers to bigrams that have at least one emote. Across the three classes ``emote only'' contributes on average 0.4493, ``emote+''  on average contributes 0.0938, while constituting 0.104 and 0.1036 respectively of the total features. Combined emote features constitute 0.5431 \footnote{The average is computed across positive, negative and neutral classifiers including emote features and emote+ features.} of the Gini feature importance to the performance of the model (using a RF classifier). This is despite being only 20.76\% of the features. This is summarized in Table \ref{tab:sum-features-contributions}.

\begin{table}[h]
\centering
\begin{adjustbox}{width=0.95\columnwidth,center}
\begin{filecontents*}{./csv/sum_features_contributions.csv}
,emote only,emote+,other
Importance Positive,0.5556,0.1092,0.3789
Importance Neutral,0.4191,0.0764,0.5318
Importance Negative,0.3733,0.0958,0.5557
Features Fraction, 0.104, 0.1036, 0.7924
\end{filecontents*}
\csvautobooktabular{./csv/sum_features_contributions.csv}
\end{adjustbox}
\caption{Feature Importance by Label and Features Fraction for each token type \label{tab:sum-features-contributions}}
\end{table}

\begin{figure}[h]
\centering
\includegraphics[width=0.95\columnwidth]{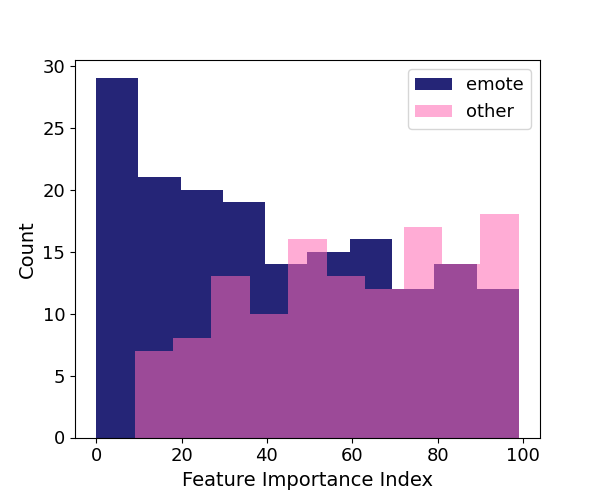}
\caption{Histograms of top 100 Gini feature importance of P1.RF.2 (for negative, positive, neutral classifiers) for emote and ``other'' features}
\label{fig:top-features-hist}
\end{figure}

We further examine the top 100 features of positive, negative and neutral classifiers. First we arrange features of each classifier by Gini index. Then we split the features into 2 categories: emote features \footnote{``emote only'' and ``emote+''} and ``other'' features. For each feature set we generate a histogram from the feature position index. Results are presented in \reffig{top-features-hist}. From the figure it is evident that emote histogram is biased toward ``0'' implying higher importance, while the ``other'' histogram is biased towards ``100''. The mean/median of the emote feature indices is 42.16/38 and the mean/median of ``other'' feature indices is 59.35/61.

To conclude, it is important to note that the performance difference between P1.RF.1 and P1.RF.2 is marginal. This implies that the introduction of bigrams is not that significant to the overall performance of the classifier. In fact the difference between a significant number of classifier combinations listed in Table \ref{tab:ec-results-accuracy} are marginal, implying that the choice of a classifier with these features is not significant, perhaps with the exception of NB. However, the presence of emote features is significant.

\section{\framework{} - Learning Out Of Vocabulary Emotions}\label{sec:framework}

Now that we have established solid baselines for the fully supervised case, we consider the task of nearing these benchmarks with a solution that resists drift and requires minimal to no supervision. This is necessary because new emotes are constantly introduced, and their usage distribution frequently changes.

Our baseline models study from Section~\ref{sec:baselines} demonstrated the critical importance of including emote information in the model. In that case, that information was explicitly encoded in per-emote features. We now want to abstract away from this requirement in order to further resist drift and ensure generalization to new emotes.

\subsection{\framework{}} We introduce a simple but powerful framework that successfully meets the requirements above. In particular, our framework---which we call \framework{}--- is able to Learn Out Of Vocabulary Emotions, and enrich existing models with this knowledge.

\begin{figure}[h]
    \centering
    \includegraphics[width=0.75\columnwidth]{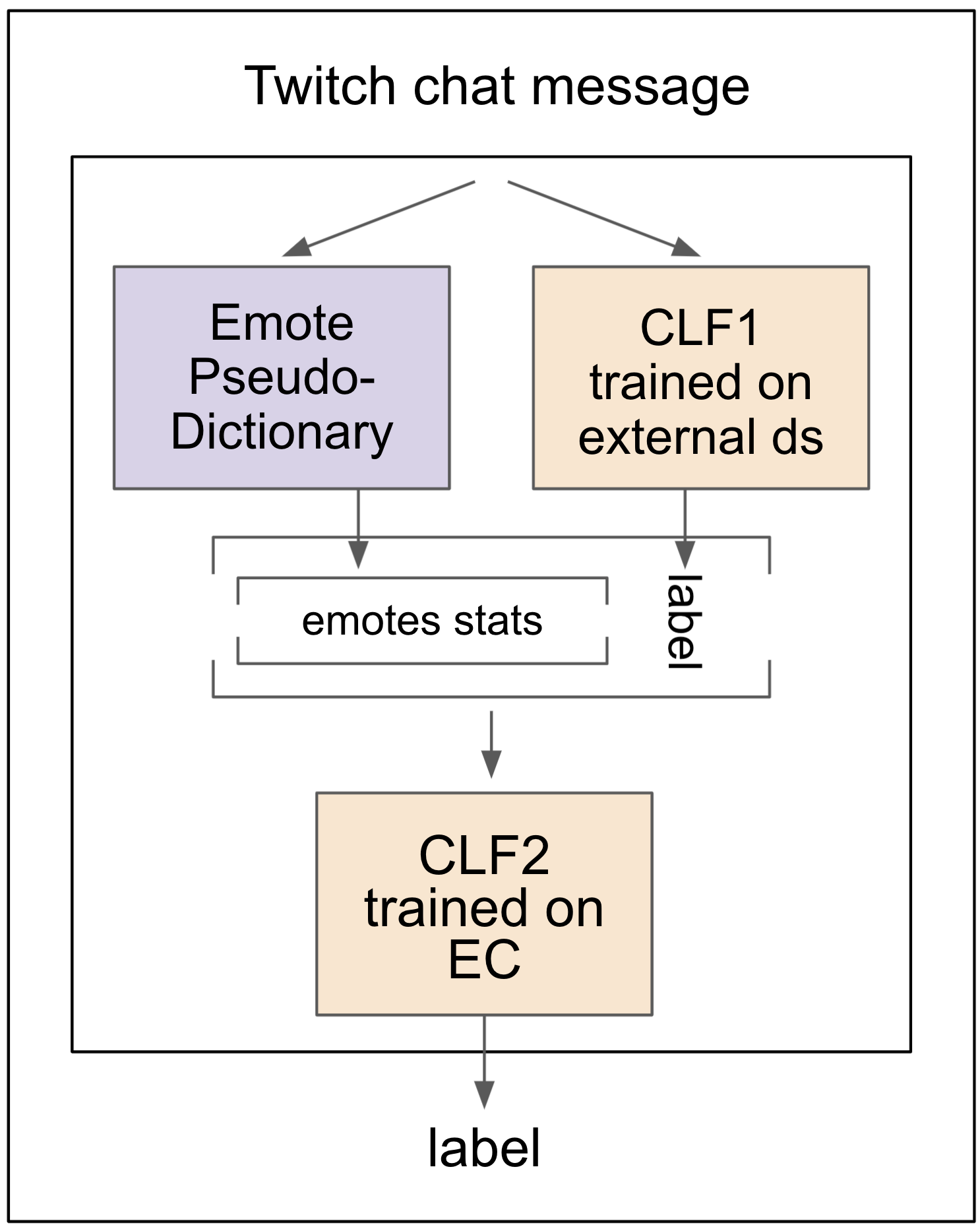}
    \caption{\framework{} framework.}
    \label{fig:tl-w2v-knn}
\end{figure}

The framework is depicted in \figurename~{\ref{fig:tl-w2v-knn}}. We start with an existing sentiment classifier: this could be a Twitch sentiment classifier that needs to be enriched with the knowledge of newly-introduced emotes; or could be a sentiment classifier trained on a completely separate dataset such as Twitter. The output of this classifier is then concatenated with emote sentiment stats obtained without needing labeled data. Specifically, for each unseen emote in the text being evaluated, its sentiment is auto-generated, averaging out the sentiment of known words in the word embedding neighborhood of the emote. Rather than introducing per-emote features, we perform ``pooling'' by keeping only a few statistics, such as the mean, max, min of the sentiment of the emotes in the text.

The \framework{} framework has several amenable properties. First, word embeddings can be trained in an entirely unsupervised manner. A periodic retraining or fine-tuning of this space removes the need to maintain a labeled dataset or a manual lexicon as new emotes are introduced. Second, our framework decouples the existing classifier from the new OOV knowledge. In practice this is very important since companies are wary of completely changing or retraining their production classifiers given they might be used across different applications. Third, while we could have encoded the emote knowledge simply by concatenating the actual emote embedding vector (with some pooling across emotes), the decision to encode just a few stats (possibly even just the average inferred sentiment) is a much more robust choice: it results in just a handful of parameters which makes tuning of the final classification extremely simple and customizable (which can be done manually or learned with very few examples); these stats are also resilient to word embedding space rotations or shifts happening upon retraining or fine-tuning. Finally, we point out that our framework is not limited to sentiment analysis, nor emotes, and can be applied to other scenarios suffering from out-of-vocabulary issues.

\subsection{\pseudodictupper{}}
We trained a \wv{} model~\cite{mikolov2013efficient} on a Twitch Unlabeled Dataset \footnote{For a detailed dataset construction refer to Appendix \ref{sec:appendix-datrasets}.}, with minimal word occurrence set at 30 and a context window of size 5. This generated 444,714 embeddings comprised of words, emotes, emojis and emoticons. Emotes represented 33.3\% of the vocabulary, words 66.2\%, and the last 0.5\% was spread across emojis and emoticons. Token statistics are displayed in \ref{tab:token_types}.

\begin{table}[h]
\centering
\begin{adjustbox}{width=0.99\columnwidth,center}
\begin{filecontents*}{./csv/df_tokens_descriptive_table.csv}
,word,emote,emoji,emoti
Unique Tokens,286795,144339,1899,240
Unique Fraction,0.6619,0.3331,0.0044,0.0006
Occurences Fraction,0.922,0.0548,0.0086,0.0022
\end{filecontents*}
\csvautobooktabular{./csv/df_tokens_descriptive_table.csv}
\end{adjustbox}
\caption{The Twitch Unlabeled Dataset token statistics \label{tab:token_types}}
\end{table}

In addition to the \wv{} model we compiled a reference sentiment table. We used the VADER lexicon~\cite{conf-icwsm-HuttoG14} augmented with an emoji/emoticon lexicon~\cite{novak_sentiment_2015} \footnote{While VADER provides a large lookup table of 7500 words and 100 emoticons, and the Emoji lookup table contains 750 emojis.}. For each emote, we generated a sentiment value by finding the top 5 neighboring words in the embedded space with an existing sentiment value in the reference sentiment table and took their mean\footnote{To avoid outliers, we limited the search of sentiment-tagged up to the 1,000th nearest neighbor. Other methods such as weighting by distance, using the median, and various outlier removal techniques were explored, but a simple average worked best.}. 

We observed 0.353 RMSE when tested against Vader's vocabulary and 0.275 RMSE accuracy when we tested against 100 manually labeled emotes provided by \cite{kobs_emote-controlled_2020}. We want to point out that this method is limited, as not every emote has neighbors that are in the reference sentiment table. Due to this limitation, we are able to generate sentiment for 22,507 emotes, even though we have embedding for over 144,000 emotes. Despite this limitation, automatically labeling over 22,000 emotes is a tremendous leap forward, as only 100 emotes have been classified before in the literature.

\subsection{External Datasets} We used the EC dataset described in Section~\ref{sec:baselines} and three publicly available datasets for ternary sentiment classification, Rotten Tomatoes (RT) \cite{pang-lee-2005-seeing}, Twitter Dataset (T) of manually labeled tweets \cite{eisner_emoji2vec_2016} and sampling Yelp Dataset \footnote{\url{https://www.yelp.com/dataset}} (Y). All datasets were split in a stratified fashion with 80\% designated for training and 20\% for testing. RT has 8,528 with 42\%/20\%/38\% positive/neutral/negative split; Twitter has 64,596 examples with 29\%/46\%/24\% class split. For Yelp we used 150,000 examples with balanced classes.

\subsection{Final Classification} For the second stage of the model, we incorporated abstracted emote information in the form of their sentiment stats and combined it with the prediction of the classifier trained on external data. The resulted feature vector is shown in column 1 of Table~\ref{table:ds-x-EC-transfer-fi}. Using these features we trained a secondary classifier using the EC dataset. Despite using EC for training, we are only effectively using this data for statistical information about emotes.

\begin{table}[h]
\centering
\begin{filecontents*}{./csv/ds_x_EC_emote_stats_accuracy.csv}
,None,ME,SVM,RF
None,,61.11,61.64,65.61
EC,71.16,71.16,71.16,71.16
RT,36.77,61.64,61.11,65.08
T,43.92,64.29,64.55,69.31
Y,41.8,61.38,62.17,65.61
\end{filecontents*}
\csvautobooktabular{./csv/ds_x_EC_emote_stats_accuracy.csv}
\caption{Accuracy results for \framework{} variants tested on EC dataset. Each row is an external dataset that CLF1 is trained on. Each column is a CLF1 variant.}
\label{table:ds-x-EC-transfer}
\end{table}

\subsection{Results} 
Accuracy results for LOOVE variants \footnote{Trained using P1 processing and tested against the EC test set} are presented in Table~\ref{table:ds-x-EC-transfer}. As depicted in \reffig{tl-w2v-knn} \framework{} is composed of 2 classifiers: CLF1 and CLF2. CLF1 is trained on an external dataset. CLF2 is trained on EC dataset using the outputs of \pseudodict{} stats and CLF1 output. The idea is to maximize the use of external datasets, while minimizing the reliance on EC. In our experiment we trained CLF1 on 4 external datasets: EC, RT, T and Y using three algorithms: ME, SVM and RF. Additionally we tested 2 ``edge cases''. For the first edge case we removed CLF1 so the model only predicts using \pseudodict{}. For the second edge case we removed \pseudodict{} and CLF2 testing the performance of CLF1.

In Table~\ref{table:ds-x-EC-transfer} each numerical column represents an algorithm choice for CLF1. Each numerical row refers to the dataset CLF1 is trained on. The first numerical column depicts the first edge case. The first numerical row of represents the second edge case.

As expected, CLF1 trained on EC gives the best performance irrespective of CLF2. However, The best performance independent of EC training for CLF1 is obtained when using RF for CLF1 trained on Twitter data (RF.T). This combination achieves 69.31\% accuracy which is on par with the fully supervised SoTA baselines obtained in Section~\ref{sec:baselines}. We want to note that CLF1 trained only on Twitter without any emote information performs worse than a coin flip when applied to Twitch data. However, when enriched by \framework{}, its performance shoots up and nearly matches our best supervised benchmark.

In Table \ref{table:ds-x-EC-transfer-fi} we examine RF.T's features for CLF2. We again see that the driving features are emotes. Only 0.0708 Gini importance comes from the predicted label of CLF1, the rest 0.9292 are driven by emotes stats.

\begin{table}[h]
\centering
\begin{filecontents*}{./csv/ds_x_EC_emote_stats_accuracy_fi.csv}
Feature Name,Importance
mean emote sentiment,0.2853
min emote sentiment,0.2568
max emote sentiment,0.2546
number of emotes,0.0968
source ds predicted label,0.0708
std emote sentiment,0.0357
\end{filecontents*}
\csvautobooktabular{./csv/ds_x_EC_emote_stats_accuracy_fi.csv}
\caption{\framework{} framework: feature importances in descending order.}
\label{table:ds-x-EC-transfer-fi}
\end{table}

\section{Word Embedding Space Analysis}\label{sec:discussion}

Given the vital importance of emotes and the success of the \framework{} we want to examine the \wv{} embedding space that \framework{} is based on and take a closer look at the structure of the \pseudodict{}.

\begin{figure}[h]
    \centering
    \includegraphics[width=\columnwidth]{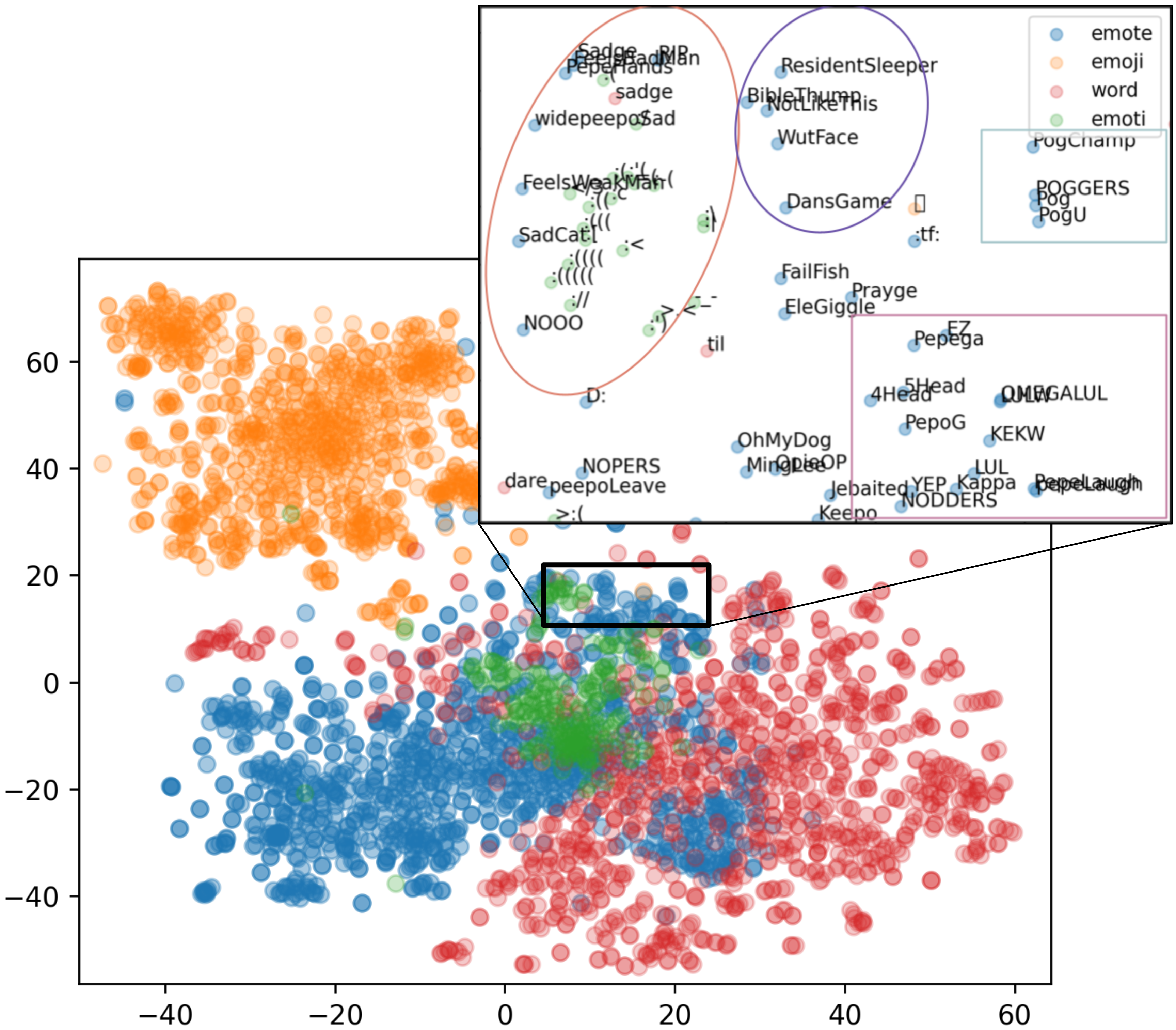}
    \caption{Top emotes, emojis, words, and emoticons (t-SNE with $perplexity=50$, $n_{iters}=3000$.). The orange oval is sadness, the purple oval is annoyance/disappointment, the pink square represents laughing/trolling, and the blue square excitement.}
    \label{fig:tsne-combined}
\end{figure}

\begin{figure}[h!]
    \centering
    \includegraphics[width=0.9\columnwidth]{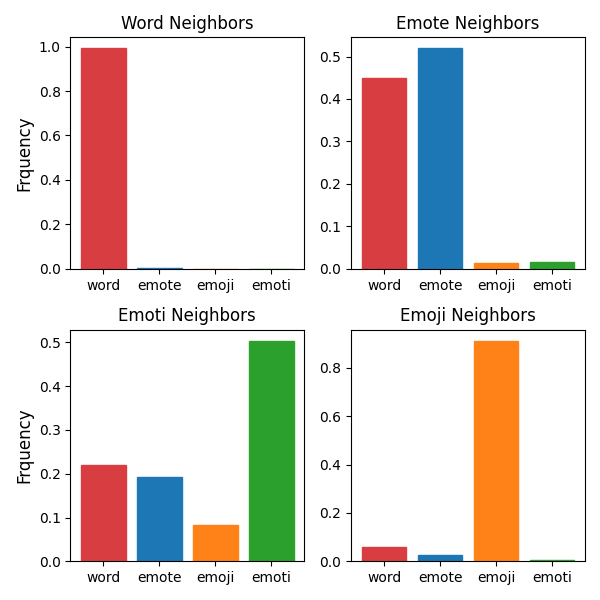}
    \caption{Distribution of neighbors for each type.}
    \label{fig:neighboring_tokens_fractions}
\end{figure}

We used t-SNE to visualize embeddings in 2D of the top 1000 emotes, 1000 words, 1000 emojis, and 240 emoticons, for a total of 3240 tokens (\figurename~{\ref{fig:tsne-combined}}). Visually, one can see that words, emotes, and emoticons overlap while the emoji cluster is more isolated. However, it is also visually evident that tokens cluster by their corresponding type. In \figurename~{\ref{fig:neighboring_tokens_fractions}}, we show the distributions of the 100 closest neighbors for each token type \footnote{looking at the same 3,240 tokens from the t-SNE visualization}. We can see that emojis indeed tend to cluster around their own type with very little exceptions, just like words do. Emotes and emoticons also like to cluster around their own type, but have neighbors from other types as well. A partial success of the \pseudodict{} can be attributed to the fact that emotes tend to cluster around words with 0.45 frequency. Since the labeled tokens from VADER are predominantly words, these are used as neighbors in $k$-NN to learn emote sentiment. This is why it is possible to construct the \pseudodict{} without relying on the sentiment of the nearest neighbor emotes themselves.

\begin{figure}[h]
    \centering
    \includegraphics[width=\columnwidth]{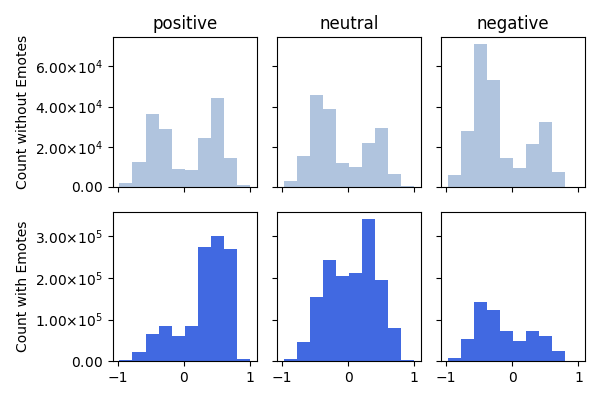}
    \caption{Top row depicts the sentiment of 1000 neighbors for each token of the original lookup sentiment table. Bottom row is for the derived emote sentiment.}
    \label{fig:hist_neighbors_by_sentiment}
\end{figure}

The box overlaying \figurename~{\ref{fig:tsne-combined}} zooms into a particular area of the space where we can find four distinct clusters representing the emotions of sadness, annoyance/disappointment, laughing/trolling, and excitement (``PogChamp''-like emotes). We examine this trend---clustering by sentiment---to see if it remains true across the embedded space. Using tokens from the reference sentiment table from Section~\ref{sec:framework}, we looked at the sentiment of the top 1,000 token neighbors and plotted sentiment histograms by label type (\figurename~{\ref{fig:hist_neighbors_by_sentiment}}, top row). Since the sentiment in the lookup table is a float between -1 (negative sentiment) and 1 (positive sentiment), we define each label type using an even 0.66 partitioning. In addition to tokens in the sentiment lookup table, we also plotted the derived emote sentiment (\figurename~{\ref{fig:hist_neighbors_by_sentiment}}, bottom row). 

The top row shows that all distributions are bimodal, with only the negative distribution being significantly biased towards the negative sentiment. On the other hand, the distributions from the bottom row are more pronounced. We can see that the positive distribution now has a clearly defined shift towards 1. The neutral distribution is a lot less bimodal. However, the negative distribution is not as pronounced as before, though it is still heavily biased towards -1. Overall, the newly generated distributions have a more pronounced bias towards the expected sentiment class. When we look at the derived emote sentiment, the bottom row of (\figurename~{\ref{fig:hist_neighbors_by_sentiment}}, we see a more prominent distribution for the positive and neutral class, with a slight bias towards negative sentiment for negative classes. This serves as an indicator that on average local neighbors tend to have the same sentiment, further illustrating the strength of our \pseudodict{}. 

In Appendix \ref{sec:appendix-emote_case_study} we present a case study involving ``FeelsGoodMan'' and ``FeelsBadMan'' emotes as an example of \pseudodict{} use for emote interpretation. In Appendix \ref{sec:appendix-embeddings} we propose several applications of the Twitch \wv{} embeddings in other fields. 

There are many problems that still need to be addressed. Despite establishing a new Twitch sentiment baseline, which performs sentiment analysis on par with other methods and datasets in terms of accuracy \cite{zimbra_state---art_2018}, overall performance can still be improved. A potential improvement to our transfer learning model as well as emote \pseudodict{} could be in the construction of an actual synonym space, rather than directly using w2v space. A notable approach to create a synonym-antonym space has been proposed in the literature by \cite{samenko_synonyms_2020}.  Using this kind of vector space to find both synonym and antonym emotes could be more successful.

\section{Conclusion}

We created multiple baselines for sentiment analysis that in the best case outperformed the previous metric by \pointsahead{} percentage points. We established the importance of emotes in sentiment analysis of Twitch data by examining the features of the baseline models, showcasing the importance of emote features. We then introduced our \framework{} unsupervised framework, which abstracts away from the explicit use of emotes as features and uses emote stats along with a classifier trained on non Twitch data to predict sentiment. This model performs nearly on par with fully supervised baselines. \framework{} is based on \wv{} embeddings trained on over 313 million Twitch chat messages in conjunction with $k$-NN. A driving feature behind the framework is a \pseudodict{}  which can be used to derive sentiment for unknown emotes. Using this \pseudodict{}, we created a sentiment table for $22,507$ emotes. This is the first case of emote understanding on this scale.

\bibliographystyle{acl_natbib}
\bibliography{anthology,custom}

\appendix

\section{Emote Case study}\label{sec:appendix-emote_case_study}
\begin{figure}[h]
    \centering
    \includegraphics[width=0.99\columnwidth]{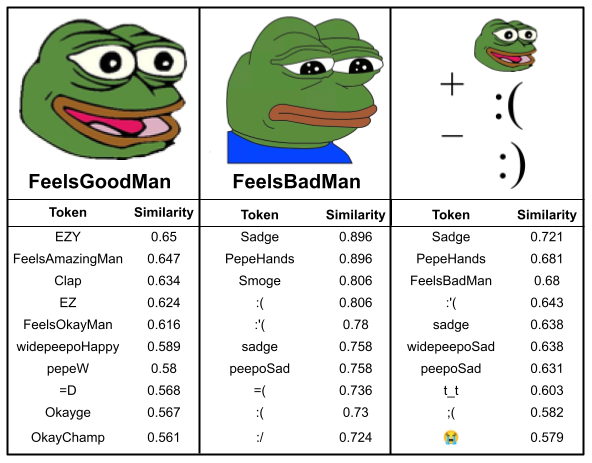}
    \caption{Neighbors of ``FeelsGoodMan'' vs. ``FeelsBadMan'' (all strings besides the emoticons above are emotes). Adding ``:('' to ``FeelsGoodMan'', subtracting ``:)''}
    \label{fig:feels-man}
\end{figure}

In \figurename~{\ref{fig:tsne-combined}} we showed a zoomed-in t-SNE plot of four distinct clusters representing the emotions of sadness, annoyance/disappointment, laughing/trolling, and excitement (``PogChamp''-like emotes). Here we present a case study, looking at two contrasting emote representatives, ``FeelsGoodMan'' and ``FeelsBadMan''. We aggregate the top 10 neighbors for each emote and display them in \figurename~{\ref{fig:feels-man}}.

From the figure it is evident that the top neighbors for each are semantically similar. The top neighbors for ``FeelsGoodMan'' are ``EZY'', ``EZ'', ``FeelsAmazingMan'', ``FeelsOkayMan'', ``widepeepoHappy'', ``pepeW'' and ``Okayge''. Each of these can be considered a different emote "flavor" of ``FeelsGoodMan'', depicting ``Pepe the Frog'' with a positive connotation. Other neighbors are used with a positive sentiment as well. 

``FeelsBadMan'' tells a similar story, but with a polar opposite sentiment. ``Sadge'', ``PepeHands'', ``Smoge'', ``sadge'', ``peepoSad'' again depict ``Pepe'' just like ``FeelsBadMan'', but with a sad, negative connotation. Other neighbors, do not feature ``Pepe'' but represent sadness as well. It is quite remarkable that the neighbors in both cases not only contain the same semantics as the original emote, but several feature ``Pepe'' as well. 

To further strengthen the case we show that it is possible to find ``FeelsBadMan'' using vector additions starting from ``FeelsGoodMan'' (and vice-versa) . As demonstrated by \cite{mikolov2013efficient} with the $Woman + King - Man = Queen$ example, we observed that by adding the frown emoticon ``:('' to ``FeelsGoodMan'' and subtracting the smile emoticon ``:)'', we obtain ``FeelsBadMan'' in the top 3 closest embeddings \figurename~{\ref{fig:feels-man}}.

\section{Twitch Unlabeled Dataset}\label{sec:appendix-datrasets}
 Our unlabeled dataset consisted of 313M chat messages from 521 thousand streams over the course of 1 week in April (06/07/21 - 06/13/21). There was an average of roughly 45K unique streamers per day. The number of messages per stream varies wildly, depending on the popularity of the streamer and the game they are playing. Up to 30\% of streams on any given day receive no messages. Looking at the data for a randomly selected day (06/08/21), the median number of messages for a stream is 117, the mean 744, with a STD of 13,585. The top 1200 streams account for 50\% of all messages, while representing 1.7\% of all streams (71,917).

\subsection{Emotes Dataset} We fetched 8.06M emotes from three sources: the Twitch official API, FrankerFaceZ (FFZ), and BetterTTV (BTTV). FFZ consists of 253,335 emotes, BTTV consists of 381,389 emotes, with the remaining Twitch official emotes. Of these, 41k emotes appear in more than one group. While the number of Twitch official emotes dwarf the other two, FFZ and BTTV emotes are incredibly popular, representing 44 of the top 100 most used emotes. 

\subsection{Trends}
\begin{figure}[h!]
    \centering
    \includegraphics[width=\columnwidth]{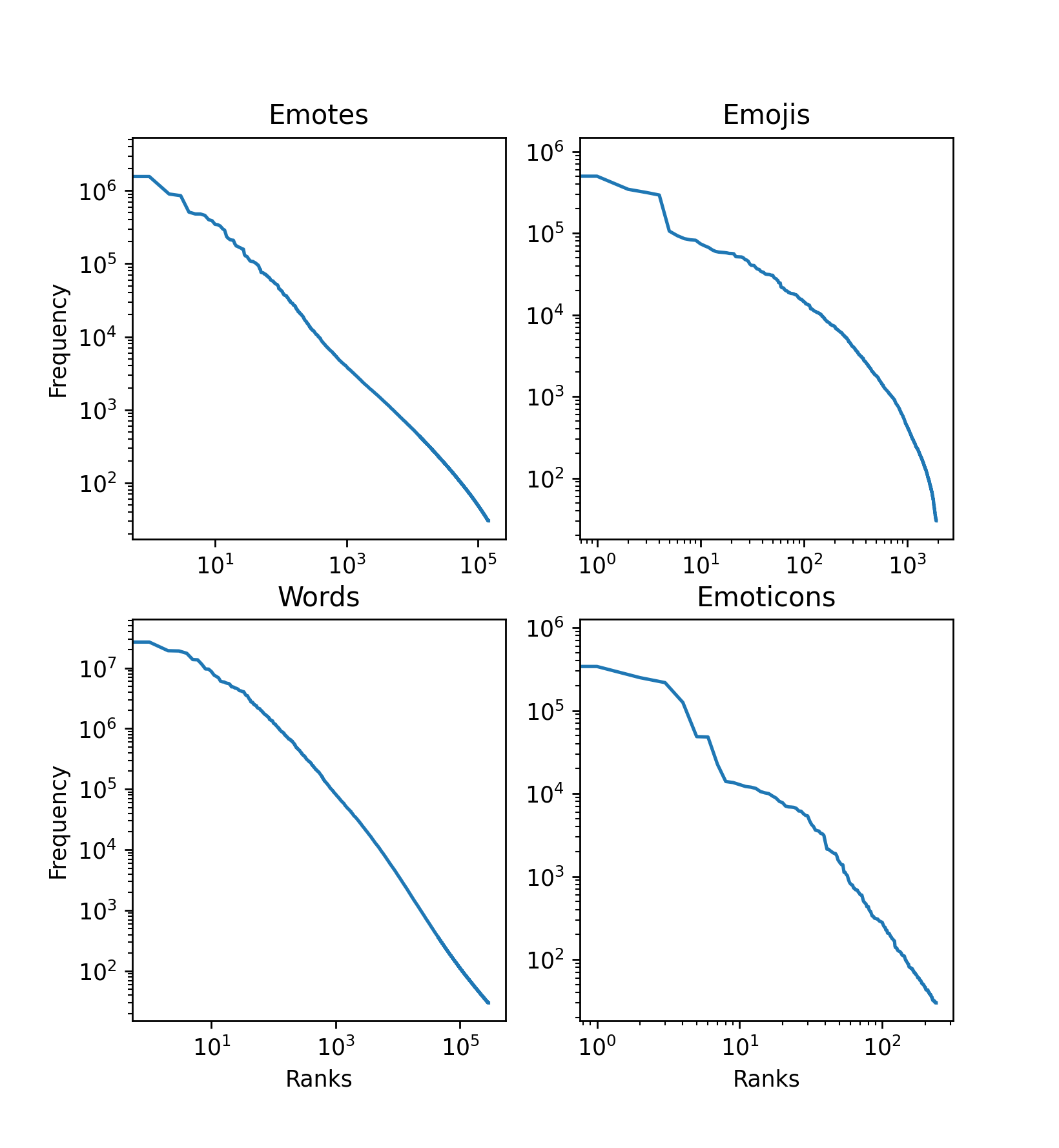}
    \caption{Frequency vs Rank by token type (log-log)}
    \label{fig:FGM}
\end{figure}

Considering that emotes account for 33\% of unique tokens in the Twitch Unlabeled Dataset (as depicted in \ref{tab:token_types}), we wanted to understand their usage frequency. By generating a rank-frequency distribution, we showed that emotes follow a power law (\figurename~{\ref{fig:FGM}}). In fact, emote rank-frequency distribution is quasi-Zipfian with a power of 0.97. Similarly, we observe that words follow a power law, as expected. However, emojis and emoticons behave somewhat differently.

\section{Embeddings Applications}\label{sec:appendix-embeddings}
In addition to using embeddings for sentiment analysis, there are other useful ways to apply these new-found embeddings. They can be used to learn slang immediately as it proliferates in the wild, improve brand safety classifiers, quickly extend a knowledge graph or add dimension to existing nodes within it, and build improved classifiers for games, genres, and industries. 

Today, slang evolves and proliferates extremely quickly due to the ubiquity of the Internet and peer-to-peer communication platforms. A platform like Twitch can be used to learn synonyms and replacements for common words and slang. The huge volume of messages and the culture of the Twitch user base makes it an ideal place to learn about new memes and slang in an unsupervised way, since they will often be some of the first users. 

Brand safety is a related application. Many moderation tools rely on keywords and regular expressions to detect profane, racist, toxic, and sexual content. While these are precise, they are likely to miss clever and new misspellings, and will certainly miss entirely new strings which are being used to “safely” convey the same meaning as their known counterparts. These could be words, emojis, or even emotes. This \wv{} model provides a way to organically learn which alternate constructions are being used to circumvent existing moderation filters.

Another application is in expanding a Knowledge Graph to incorporate related entities, or to add variations/nicknames of known entities. This works best in the gaming-space since that is the community’s focus, but also for television shows, cryptocurrencies, sports, etc. Given the string ``hikaru'', the closest embeddings are names of dozens of other chess players. In \ref{table:df_words_and_similarities} Given the string ``morde'' (short for ``Mordekaiser'', a champion in League of Legends), the model returns other champions from the game and their nicknames. This was also tested for a few other words such as ``vaxx'' (short for vaccine), ``grau'', a popular gun in the game Call of Duty: Warzone, and other words. 

\begin{table}
\centering
\begin{adjustbox}{width=\columnwidth,center}
\begin{filecontents*}{./csv/dataframe_entities_and_similarities.csv}
,hikaru,morde,grau,vaxx,troll
1,danya,darius,ffar,vax,bully
2,levy,vayne,kar98,vacc,tryhard
3,kasparov,aatrox,mac10,vacine,bait
4,anish,panth,spr,vaccin,jebait
5,naroditsky,leona,bruen,vaccination,grief
6,samay,voli,bullfrog,phizer,ban
7,lefong,trundle,m13,vaccine,simp
8,fabi,garen,ram7,astrazeneca,toxic
9,nihal,heimer,dmr,moderna,bm
10,nili,sion,fara,covid-19,trolling
\end{filecontents*}
\csvautobooktabular{./csv/dataframe_entities_and_similarities.csv}
\end{adjustbox}
\caption{Five words and their most similar tokens.}
\label{table:df_words_and_similarities}
\end{table}


\end{document}